%% file: main.tex
\documentclass[runningheads]{llncs}
\usepackage[T1]{fontenc}
\usepackage{graphicx,verbatim}
\usepackage{amsmath,amssymb}
\usepackage{longtable}
\usepackage{pdflscape}
\usepackage{booktabs}
\usepackage{multirow}
\usepackage{siunitx}
\usepackage{hyperref}

\newcommand{\ourmodel}{MINT}  

\begin{document}
\title{\ourmodel: Molecularly Informed Training with Spatial Transcriptomics Supervision for Pathology Foundation Models}
\titlerunning{MINT: Molecularly Informed Training with
ST Supervision}
\begin{comment}  %% Removed for anonymized MICCAI submission
\author{First Author\inst{1}\orcidID{0000-1111-2222-3333} \and
Second Author\inst{2,3}\orcidID{1111-2222-3333-4444} \and
Third Author\inst{3}\orcidID{2222--3333-4444-5555}}
%
\authorrunning{F. Author et al.}
\institute{Princeton University, Princeton NJ 08544, USA \and
Springer Heidelberg, Tiergartenstr. 17, 69121 Heidelberg, Germany
\email{lncs@springer.com}\\
\url{http://www.springer.com/gp/computer-science/lncs} \and
ABC Institute, Rupert-Karls-University Heidelberg, Heidelberg, Germany\\
\email{\{abc,lncs\}@uni-heidelberg.de}}

\end{comment}

\author{Minsoo Lee \and
Jonghyun Kim \and
Juseung Yun \and Sunwoo Yu \and Jongseong Jang}
\authorrunning{Lee et al.}
\institute{LG AI Research \\
\email{\{minsoo.lee, kimjh0107, js.yun, ysw0920, j.jang\}@lgresearch.ai}}

% \author{Anonymized Authors}  %% Added for anonymized MICCAI submission
% \authorrunning{Anonymized Author et al.}
% \institute{Anonymized Affiliations \\
%     \email{email@anonymized.com}}
  
\maketitle              % typeset the header of the contribution
%

%%%%%%%%% ABSTRACT
\input{section/0_abstract}

%
%
%
%%%%%%%%% Introduction
\input{section/1_intro}

%%%%%%%%% Method
\input{section/3_method}

%%%%%%%%% Experiment
\input{section/4_experiment}
\input{section/5_conclusion}


\end{document}

%% file: section/0_abstract.tex
\begin{abstract}
Pathology foundation models learn morphological representations through self-supervised pretraining on large-scale whole-slide images, yet they do not explicitly capture the underlying molecular state of the tissue. Spatial transcriptomics technologies bridge this gap by measuring gene expression in situ, offering a natural cross-modal supervisory signal. We propose MINT (Molecularly Informed Training), a fine-tuning framework that incorporates spatial transcriptomics supervision into pretrained pathology Vision Transformers. MINT appends a learnable ST token to the ViT input to encode transcriptomic information separately from the morphological CLS token, preventing catastrophic forgetting through DINO self-distillation and explicit feature anchoring to the frozen pretrained encoder. Gene expression regression at both spot-level (Visium) and patch-level (Xenium) resolutions provides complementary supervision across spatial scales. Trained on 577 publicly available HEST samples, MINT achieves the best overall performance on both HEST-Bench for gene expression prediction (mean Pearson $r{=}0.440$) and EVA for general pathology tasks ($0.803$), demonstrating that spatial transcriptomics supervision complements morphology-centric self-supervised pretraining.

% Authors must provide keywords and are not allowed to remove this Keyword section.

\end{abstract}

%% file: section/1_intro.tex
\section{Introduction}

Foundation models pretrained on large-scale whole-slide image (WSI) collections have become central to computational pathology~\cite{hoptimus,uni,virchow,conch}. Through self-supervised objectives such as DINO~\cite{dino} and DINOv2~\cite{dinov2}, these models learn morphological representations from millions of histopathology tiles without manual annotation, enabling effective transfer to diverse downstream tasks including tissue classification, biomarker prediction, and survival analysis. Recent scaling efforts---H-optimus-0~\cite{hoptimus} (ViT-G on 500K slides), UNI2-h~\cite{uni2} (ViT-H on 350K slides), and Virchow2~\cite{virchow2} (ViT-H on 3.1M slides)---have progressively improved performance across standardized benchmarks.

However, these models are trained \emph{exclusively} on visual patterns. Histopathology images implicitly encode the molecular state of the tissue microenvironment: cellular composition, signaling pathway activity, and gene expression programs all manifest in tissue morphology. Spatial transcriptomics (ST) technologies provide direct measurements of this link. Spot-level platforms~\cite{visium}, from the original Spatial Transcriptomics to its commercial successor Visium (10x Genomics), profile transcriptomes at spot-level resolution ($\sim$55\,$\mu$m, roughly 10--50 cells per spot), while Xenium~\cite{xenium} provides single-molecule transcript detection at sub-cellular resolution. Recently, the HEST dataset~\cite{hest} has aggregated over 1,100 such paired histology--transcriptomics samples spanning 131 organs, making large-scale training data for this modality publicly available.

A natural question arises: \emph{can spatially-resolved gene expression serve as supervisory signal to improve the representations of pretrained pathology foundation models?} Prior work has explored gene expression prediction as an independent supervised task~\cite{stnet,histogene,bleep,thitogene}, but has not used it to enhance foundation model representations. The key challenge is that fine-tuning on gene expression objectives risks catastrophic forgetting of the morphological representations acquired during large-scale pretraining.

We introduce MINT (Molecularly Informed Training with Spatial Transcriptomics Supervision), a multi-task fine-tuning framework that addresses this challenge through three design principles. First, rather than modifying the CLS token that encodes morphological features, we append a dedicated learnable \emph{ST token} to the ViT input sequence, providing the model with a separate channel for molecular information. Second, we combine DINO self-distillation for continued visual learning with explicit feature distillation from the frozen pretrained model, creating a dual mechanism against catastrophic forgetting. Third, we exploit both spot-level and Xenium patch-level gene expression as complementary supervision at different spatial resolutions.

Our contributions are as follows:
\begin{enumerate}
    \item We propose MINT that incorporates spatial transcriptomics supervision into pretrained pathology ViTs while preventing catastrophic forgetting through a dedicated ST token and dual distillation mechanisms.
    \item We show that the ST and CLS tokens capture complementary information---the ST token specializes for molecular signals while the CLS token retains morphological transferability---and that combining both yields consistent improvements in a backbone-agnostic manner.
    \item Under the official HEST-Bench and EVA evaluation framework, MINT achieves the best overall performance on both benchmarks (mean Pearson $r{=}0.440$ and EVA average $0.803$).

\end{enumerate}

%% file: section/3_method.tex
\section{Method}
\subsection{Overview}
\begin{figure}[t]
  \centering
  \includegraphics[width=\textwidth]{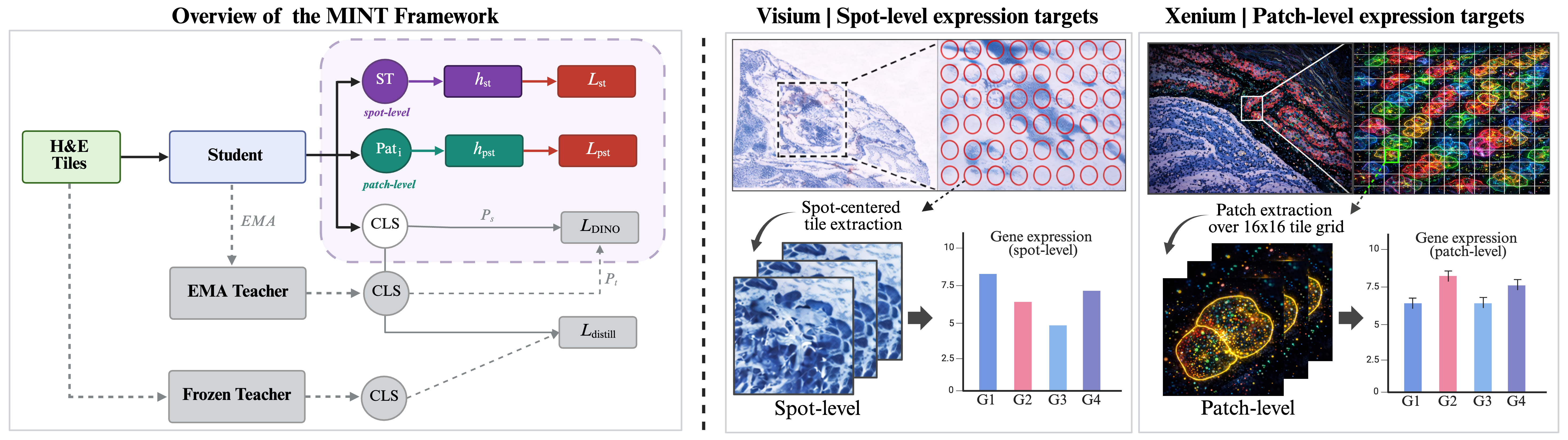}
    \caption{MINT framework. The student ViT, augmented with a learnable ST token, outputs CLS, ST, and patch token representations. Gene expression regression from the ST token ($\mathcal{L}_{\text{ST}}$) and patch tokens ($\mathcal{L}_{\text{pST}}$) provides transcriptomic supervision, while DINO self-distillation ($\mathcal{L}_{\text{DINO}}$) and feature anchoring ($\mathcal{L}_{\text{distill}}$) preserve morphological representations. Only the student is updated via backpropagation.}
  \label{fig:architecture}
  \end{figure}

Given a pretrained ViT encoder $f_\theta$ with embedding dimension $D$, MINT constructs a student--teacher framework and fine-tunes with four complementary objectives (Fig.~\ref{fig:architecture}). The student and teacher are initialized from $f_\theta$; the student is updated via gradient descent while the teacher tracks the student through exponential moving average (EMA). A frozen copy of $f_\theta$ serves as a distillation anchor. The key idea is to augment the ViT with a learnable \emph{ST token} that encodes transcriptomic information, while the original CLS and patch tokens retain their morphological roles.

\subsection{ST Token Design}
\label{sec:st_token}

We augment the ViT input with a learnable \emph{ST token} $\mathbf{t}^{\text{st}} \in \mathbb{R}^D$, appended alongside the CLS token and $N$ patch tokens (with $N{=}256$ for ViT-g/14 at $224{\times}224$). After processing through all transformer layers, the model produces three output representations: the CLS token $\mathbf{z}^{\text{cls}} \in \mathbb{R}^D$ retaining morphological features, the ST token $\mathbf{z}^{\text{st}} \in \mathbb{R}^D$ encoding transcriptomic features, and patch tokens $\{\mathbf{z}_i^{\text{pat}}\}_{i=1}^N$ providing spatially-localized representations.

The motivation for a separate token, rather than supervising the CLS token itself for gene expression, is twofold. First, it preserves the CLS token's morphological representations acquired through large-scale WSI pretraining. Directly supervising the CLS token for gene expression risks overwriting these features; the separate ST token instead specializes for gene expression while the CLS token continues to serve its original role in the self-supervised and feature anchoring objectives. Second, the ST token participates in self-attention with CLS and patch tokens across all transformer layers, enabling it to learn molecular representations from the full spatial context. At inference, concatenating both tokens ($[\mathbf{z}^{\text{cls}} \| \mathbf{z}^{\text{st}}] \in \mathbb{R}^{2D}$) combines both types of information, while using CLS alone recovers the original feature space.

\subsection{Training Objectives}
\label{sec:losses}

\subsubsection{DINO Self-Distillation.}
Following DINO~\cite{dino}, we apply multi-crop augmentation: 2 global crops ($224{\times}224$) and 8 local crops ($96{\times}96$). The student processes all 10 views while the teacher sees only global views. Both project their CLS tokens through a shared MLP head to a $65{,}536$-dimensional probability space with centering and sharpening:
\begin{equation}
    \mathcal{L}_{\text{DINO}} = -\sum_{x \in \mathcal{V}_g} \sum_{\substack{x' \in \mathcal{V} \\ x' \neq x}} P_t(x) \log P_s(x')
\end{equation}
where $\mathcal{V}_g$ and $\mathcal{V}$ denote global and all crops, and $P_t$, $P_s$ are teacher and student distributions. This loss maintains the self-supervised learning dynamics of the original pretraining, enabling continued visual representation learning during fine-tuning.

\subsubsection{Feature Distillation.}
To explicitly anchor the student's CLS representation near the pretrained feature space, we minimize the $L_2$ distance to the frozen model's output on all crops:
\begin{equation}
    \mathcal{L}_{\text{distill}} = \frac{1}{|\mathcal{V}|} \sum_{x \in \mathcal{V}} \left\| \mathbf{z}^{\text{cls}}_{\text{student}}(x) - \mathbf{z}^{\text{cls}}_{\text{frozen}}(x) \right\|_2^2
\end{equation}
This provides a direct regularization against catastrophic forgetting, complementing the implicit regularization from DINO's EMA teacher. The combination of both mechanisms---EMA-based self-distillation and explicit feature anchoring---provides redundant safeguards for representation stability.

\subsubsection{Spot-Level ST Regression.}
Each training tile corresponds to a Visium spot. We define a shared vocabulary of $G{=}38{,}710$ genes as the union across all slides; since each slide measures only a subset, the loss is computed only over genes present in the current slide. The ST token is passed through an MLP head $h_{\text{st}}: \mathbb{R}^D \to \mathbb{R}^G$ to predict $\log(1{+}x)$-transformed expression values. To further focus on informative genes, we apply stochastic highly variable gene (HVG) selection~\cite{hvg}: at each iteration, with probability $p_{\text{hvg}}$ we restrict the loss to the top-$k$ HVGs, and otherwise use all measured genes:
\begin{equation}
    \mathcal{L}_{\text{ST}} = \frac{1}{|\mathcal{G}_s|} \sum_{g \in \mathcal{G}_s} \left( h_{\text{st}}(\mathbf{z}^{\text{st}})_g - y_g \right)^2
\end{equation}
where $\mathcal{G}_s$ is the (stochastically) selected gene subset and $y_g$ the target expression.

\subsubsection{Patch-Level Xenium Regression.}
Spot-level platforms provide one expression profile per spot, averaging over 10--50 cells. Xenium technology~\cite{xenium}, in contrast, detects individual transcripts at sub-cellular resolution, enabling construction of patch-level targets. For each $16{\times}16$ grid of ViT patches within a tile, we aggregate Xenium transcripts falling within each patch's spatial extent to create per-patch expression vectors.

Each patch token $\mathbf{z}_i^{\text{pat}}$ is passed through a dedicated MLP head $h_{\text{pst}}: \mathbb{R}^D \to \mathbb{R}^{G_{\text{xen}}}$, where $G_{\text{xen}} = 5{,}783$ is the union of genes across available Xenium panels. Since each panel covers a different gene subset, the loss for a given sample is restricted to its panel's genes. We further apply \emph{positive-only} supervision: the loss is computed only over patches containing at least one detected transcript, avoiding penalization for regions without detected transcripts:
\begin{equation}
    \mathcal{L}_{\text{pST}} = \frac{1}{|\mathcal{P}^+|} \sum_{p \in \mathcal{P}^+} \frac{1}{|\mathcal{G}|} \sum_{g \in \mathcal{G}} \left( h_{\text{pst}}(\mathbf{z}_p^{\text{pat}})_g - y_{p,g} \right)^2
\end{equation}
where $\mathcal{P}^+$ is the set of patches containing at least one detected transcript and $\mathcal{G}$ the set of genes measured by the sample's Xenium panel.

\subsubsection{Total Objective.}
The four losses are combined as:
\begin{equation}
    \mathcal{L} = \mathcal{L}_{\text{DINO}} + \lambda_{\text{distill}}\,\mathcal{L}_{\text{distill}} + \lambda_{\text{ST}}\,\mathcal{L}_{\text{ST}} + \lambda_{\text{pST}}\,\mathcal{L}_{\text{pST}}
    \label{eq:total}
\end{equation}
where $\lambda_{\text{distill}}$, $\lambda_{\text{ST}}$, and $\lambda_{\text{pST}}$ are scalar weights for each auxiliary objective.

%% file: section/4_experiment.tex
\section{Experiments}
\subsection{Training Setup}
\label{sec:training_setup}
We fine-tune H-optimus-0~\cite{hoptimus} (ViT-g/14, 1.1B parameters) on the HEST dataset~\cite{hest}, which provides paired histology tiles and spot-level spatial transcriptomics measurements (Visium and earlier ST platforms) across diverse tissue types. From the 649 human samples in HEST, we exclude 72 samples included in HEST-Bench to prevent data leakage, yielding 577 training samples. For patch-level Xenium supervision (\S\ref{sec:losses}), we use 44 Xenium-profiled samples from HEST, oversampled $5{\times}$ to balance against the more abundant spot-level data.
We train for 90K iterations using AdamW with learning rate $5{\times}10^{-5}$, cosine schedule, batch size 64 per GPU across 8 GPUs, and bf16 mixed precision. The loss weights $\lambda_{\text{distill}}$, $\lambda_{\text{ST}}$, and $\lambda_{\text{pST}}$ in Eq.~\ref{eq:total} are each set to 100. For HVG selection, we use $k{=}200$ and $p_{\text{hvg}}{=}0.5$. We denote the resulting model as \textbf{MINT}.
At inference, we concatenate the CLS and ST tokens to form $[\mathbf{z}^{\text{cls}} \| \mathbf{z}^{\text{st}}] \in \mathbb{R}^{3072}$ as the default feature representation.

\begin{table}[t]
\centering
\small
\caption{HEST-Bench results (PCA+Ridge with 256 PCA components; Pearson correlation; higher is better). Best in \textbf{bold}, second-best \underline{underlined}.}
\label{tab:hestbench}
\resizebox{\linewidth}{!}{
\begin{tabular}{lcccccccccc}
\hline
Model & IDC & PRAD & PAAD & SKCM & COAD & READ & ccRCC & LUAD & LYMPH & Avg \\
\hline
% ResNet50~\cite{resnet} & 0.474 & 0.308 & 0.389 & 0.482 & 0.253 & 0.081 & 0.223 & 0.492 & 0.232 & 0.326 \\
CTransPath~\cite{ctranspath} & 0.511 & 0.343 & 0.438 & 0.511 & 0.229 & 0.110 & 0.228 & 0.499 & 0.235 & 0.345 \\
Phikon~\cite{phikon} & 0.533 & 0.342 & 0.443 & 0.536 & 0.259 & 0.152 & 0.242 & 0.547 & 0.237 & 0.366 \\
CONCH~\cite{conch} & 0.536 & 0.355 & 0.448 & 0.579 & 0.253 & 0.167 & 0.218 & 0.531 & 0.251 & 0.371 \\
REMEDIS~\cite{remedis} & 0.529 & 0.347 & 0.464 & 0.582 & 0.286 & 0.115 & 0.265 & 0.534 & 0.247 & 0.374 \\
CONCH 1.5~\cite{conch} & 0.544 & 0.381 & 0.457 & 0.552 & 0.280 & 0.160 & 0.218 & 0.551 & 0.270 & 0.379 \\
GigaPath~\cite{prov} & 0.551 & 0.371 & 0.477 & 0.554 & 0.301 & 0.186 & 0.239 & 0.540 & 0.249 & 0.385 \\
UNI~\cite{uni} & 0.570 & 0.314 & 0.476 & 0.625 & 0.263 & 0.176 & 0.243 & 0.551 & 0.257 & 0.386 \\
H0-mini~\cite{h0mini} & 0.586 & 0.368 & 0.492 & 0.601 & 0.249 & 0.186 & 0.267 & 0.548 & 0.263 & 0.396 \\
Virchow~\cite{virchow} & 0.570 & 0.331 & 0.488 & 0.609 & \underline{0.311} & 0.202 & 0.264 & 0.546 & 0.259 & 0.398 \\
Virchow2~\cite{virchow2} & 0.592 & 0.347 & 0.466 & 0.617 & 0.258 & 0.208 & \underline{0.279} & \underline{0.561} & 0.258 & 0.398 \\
% UNIv1.5~\cite{uni2} & \underline{0.599} & 0.365 & 0.490 & 0.640 & 0.293 & 0.224 & 0.252 & 0.559 & 0.260 & 0.409 \\
UNI2-h~\cite{uni2} & 0.590 & 0.357 & \underline{0.500} & \underline{0.660} & 0.301 & 0.223 & 0.264 & 0.558 & \underline{0.273} & 0.414 \\
H-optimus-0~\cite{hoptimus} & 0.598 & \underline{0.385} & 0.493 & 0.643 & 0.299 & \underline{0.229} & 0.265 & 0.558 & 0.260 & \underline{0.415} \\
\hline
% \textbf{MINT (Uni2-h) } ~& \textbf{0.646} & \textbf{0.386} & \textbf{0.511} & \textbf{0.695} & \textbf{0.333} & \textbf{0.232} & \textbf{0.252} & \textbf{0.597} & \textbf{0.266} & \textbf{0.435} \\
\textbf{MINT} & \textbf{0.655} & \textbf{0.389} & \textbf{0.509} & \textbf{0.695} & \textbf{0.314} & \textbf{0.244} & \textbf{0.283} & \textbf{0.597} & \textbf{0.275} & \textbf{0.440} \\
\hline
\end{tabular}}
\end{table}

\begin{table}[t]
\centering
\small
\caption{EVA evaluation framework results (higher is better). Best in \textbf{bold}, second-best \underline{underlined}. Average is the arithmetic mean over 9 benchmarks.}
\label{tab:eva}
\resizebox{\linewidth}{!}{
\begin{tabular}{lcccccccccc}
\hline
Model & BHis & CRC & Glea & MHIST & PCam & Cam16 & PANDA & CoNS & MoNuS & Average\\
\hline
Phikon-v2~\cite{phikonv2} &
0.713 & 0.939 & 0.757 & 0.777 & 0.894 & 0.805 & 0.625 & 0.627 & 0.644 & 0.753 \\
Phikon~\cite{phikon} &
0.717 & 0.940 & 0.729 & 0.803 & 0.920 & 0.798 & 0.644 & 0.627 & 0.635 & 0.757 \\
CONCH~\cite{conch} &
0.681 & 0.953 & 0.729 & 0.832 & 0.922 & 0.841 & 0.628 & 0.618 & 0.632 & 0.760 \\
hibou-L~\cite{hibou} &
0.735 & 0.932 & 0.764 & 0.839 & \textbf{0.955} & 0.823 & 0.634 & \textbf{0.646} & 0.669 & 0.777 \\
UNI~\cite{uni} &
0.785 & 0.944 & 0.750 & \underline{0.843} & 0.937 & 0.833 & 0.659 & 0.628 & 0.659 & 0.782 \\
Pv-GigaPath~\cite{prov} &
0.827 & 0.951 & 0.724 & 0.829 & 0.945 & 0.815 & 0.653 & 0.626 & \underline{0.680} & 0.783 \\
Midnight-12k~\cite{midnight} &
0.819 & \underline{0.966} & \textbf{0.800} & 0.804 & 0.929 & 0.849 & 0.649 & 0.623 & 0.659 & 0.789 \\
H-optimus-0~\cite{hoptimus} &
0.801 & 0.955 & 0.770 & \underline{0.843} & 0.943 & 0.827 & \underline{0.671} & \underline{0.644} & \textbf{0.685} & 0.793 \\
UNI2-h~\cite{uni2} &
\underline{0.859} & 0.965 & 0.775 & 0.824 & \underline{0.950} & 0.849 & 0.657 & 0.630 & 0.642 & 0.795 \\
Virchow2~\cite{virchow2} &
0.821 & \textbf{0.967} & \underline{0.783} & \textbf{0.861} & 0.938 & \textbf{0.861} & 0.646 & 0.640 & 0.669 & \underline{0.798} \\
\hline
\textbf{MINT} &
\textbf{0.864} & 0.960 & 0.769 & 0.834 & 0.943 & \underline{0.851} & \textbf{0.673} & \textbf{0.646} & \textbf{0.685} & \textbf{0.803} \\
\hline
\end{tabular}}
\end{table}

\subsection{Benchmarks and Protocols}
We evaluate on two complementary benchmarks, strictly following each benchmark's official protocol.
First, we report results on HEST-Bench, which assesses how well histology features support gene expression prediction across 9 cancer types. Following the benchmark specification, we use the standard PCA+Ridge evaluation pipeline with 256 PCA components and report Pearson correlation (higher is better).
Second, we evaluate general-purpose pathology transferability using the EVA evaluation framework, which consists of nine pathology benchmarks spanning classification (BreakHis, CRC, Gleason, MHIST, PCam), weakly-supervised whole-slide tasks (Cam16, PANDA), and nuclei segmentation (CoNSeP, MoNuSAC); we refer to~\cite{eva} for full protocol details. 
For classification and whole-slide tasks, we use the [CLS $\|$ ST] concat feature. 
For segmentation tasks, we follow the EVA protocol using spatially arranged patch features concatenated with the input image under online linear evaluation, without using the ST token.
For fair comparison, all baseline numbers reported in Tables~\ref{tab:hestbench} and~\ref{tab:eva} (except MINT) are taken from the official benchmark leaderboards.
We additionally run the official evaluation pipelines to verify that these reported baseline results are reproducible under the benchmark settings.
All MINT results are obtained using the same official evaluation pipelines without any modifications.

\subsection{Comparison with Existing Models}
Table~\ref{tab:hestbench} presents HEST-Bench results. MINT achieves the best overall mean Pearson correlation (0.440), surpassing all compared models including H-optimus-0 (0.415) and UNI2-h (0.414), and ranking first across all 9 cancer types. On EVA (Table~\ref{tab:eva}), MINT also attains the highest average (0.803), exceeding Virchow2 (0.798) and H-optimus-0 (0.793). Compared to the pretrained H-optimus-0 backbone, MINT maintains comparable or improved performance across all EVA tasks except MHIST, where a marginal decrease is observed.

Notably, MINT ranks first on both benchmarks simultaneously. HEST-Bench measures molecular prediction capability while EVA assesses general-purpose pathology transferability---two axes that are not inherently aligned. The fact that MINT improves on both indicates that spatial transcriptomics supervision provides complementary information to morphology-centric pretraining, rather than introducing a trade-off between molecular and morphological performance.

These gains are achieved not by scaling up image data, but by introducing a new supervisory modality: MINT fine-tunes on only 577 paired histology--transcriptomics samples from the publicly available HEST dataset, yet improves both molecular and morphological benchmarks. This suggests that cross-modal supervision offers a distinct and complementary axis for improving pathology foundation models beyond image-only self-supervised scaling, and that further growth of paired histology--transcriptomics datasets may yield additional gains.

\subsection{Representation Analysis}
Table~\ref{tab:repr_variants_backbone_with_base} examines how the choice of feature representation affects downstream performance across two backbone architectures.
Since the EVA segmentation benchmarks (CoNSeP, MoNuSAC) rely on spatially arranged patch embeddings rather than token-level representations, the EVA average here is computed over classification and whole-slide tasks only.
The CLS and ST tokens exhibit clear specialization: ST-only outperforms CLS-only on HEST-Bench (0.428 vs.\ 0.413 for H-optimus-0), while CLS-only outperforms ST-only on EVA (0.828 vs.\ 0.823), indicating that each token captures distinct information---molecular and morphological, respectively.
Combining both tokens improves performance on both benchmarks. Notably, element-wise summation (CLS $\oplus$ ST) uses the same dimensionality as CLS-only (1536-d) yet substantially outperforms it (e.g., 0.431 vs.\ 0.413 on HEST-Bench), confirming that the gains arise from genuinely complementary information rather than increased feature dimensionality. Concatenation ([CLS $\|$ ST]) achieves the best results across both benchmarks.
These patterns are consistently observed across both H-optimus-0 and UNI2-h backbones, confirming that the benefit of the ST token is backbone-agnostic.

To validate the necessity of the dedicated ST token, we ablate it by applying $\mathcal{L}_{\text{ST}}$ directly to the CLS token, with and without feature distillation ($\mathcal{L}_{\text{distill}}$). Both variants achieve higher HEST-Bench scores than the pretrained baseline, indicating that the CLS token successfully learns molecular information regardless of distillation. However, without distillation, EVA-CLS drops substantially (0.830$\to$0.811 for H-optimus-0; 0.840$\to$0.824 for UNI2-h), revealing catastrophic forgetting of morphological representations. Adding distillation largely recovers EVA performance (to 0.827 and 0.836, respectively), confirming its role in preventing forgetting. Nevertheless, both $\mathcal{L}_{\text{ST}}$ on [CLS] variants fall short of MINT's token-separated design on both benchmarks for both backbones: even CLS $\oplus$ ST summation outperforms the best $\mathcal{L}_{\text{ST}}$ on [CLS] variant (0.431 / 0.837 vs.\ 0.426 / 0.827 for H-optimus-0; 0.433 / 0.844 vs.\ 0.426 / 0.836 for UNI2-h), demonstrating that decoupling molecular and morphological learning into separate tokens is more effective than encoding both in a shared representation.
MINT's dedicated ST token avoids this trade-off entirely: the CLS-only variant of MINT closely matches the pretrained baseline on EVA (0.828 vs.\ 0.830 for H-optimus-0; 0.839 vs.\ 0.840 for UNI2-h), confirming that the ST token successfully decouples molecular learning from the pretrained feature space.

\begin{table}[t]
\centering
\small
\caption{Representation variants across backbones, including pretrained baselines and ablations. HEST-Bench reports mean Pearson correlation; EVA-CLS averages over classification and whole-slide tasks (7 benchmarks). Best in \textbf{bold}, second-best \underline{underlined}.}
\label{tab:repr_variants_backbone_with_base}
\renewcommand{\arraystretch}{1.05}
\begin{tabular*}{\linewidth}{@{\extracolsep{\fill}} l l l c c}
\toprule
Backbone & Training & Representation & HEST-Bench & EVA-CLS \\
\midrule

\multirow{7}{*}{H-opt-0}
& No training & CLS-only & 0.415 & 0.830 \\
\cmidrule(lr){2-5}
& $L_{\mathrm{ST}}$ on [CLS] w/o $L_{\mathrm{distill}}$ & CLS-only & 0.424 & 0.811 \\
& $L_{\mathrm{ST}}$ on [CLS] + $L_{\mathrm{distill}}$   & CLS-only & 0.426 & 0.827 \\
\cmidrule(lr){2-5}
& \multirow{4}{*}{MINT} & CLS-only               & 0.413 & 0.828 \\
&                       & ST-only                & 0.428 & 0.823 \\
&                       & CLS $\oplus$ ST (sum)  & \underline{0.431} & \underline{0.837} \\
&                       & [CLS $\|$ ST] (concat) & \textbf{0.440} & \textbf{0.842} \\

% double rule separator between backbones
\midrule
\addlinespace[-0.35em]
\midrule

\multirow{7}{*}{UNI2-h}
& No training & CLS-only & 0.414 & 0.840 \\
\cmidrule(lr){2-5}
& $L_{\mathrm{ST}}$ on [CLS] w/o $L_{\mathrm{distill}}$ & CLS-only & 0.420 & 0.824 \\
& $L_{\mathrm{ST}}$ on [CLS] + $L_{\mathrm{distill}}$   & CLS-only & 0.426 & 0.836 \\
\cmidrule(lr){2-5}
& \multirow{4}{*}{MINT} & CLS-only               & 0.410 & 0.839 \\
&                       & ST-only                & 0.426 & 0.834 \\
&                       & CLS $\oplus$ ST (sum)  & \underline{0.433} & \underline{0.844} \\
&                       & [CLS $\|$ ST] (concat) & \textbf{0.435} & \textbf{0.848} \\
\bottomrule
\end{tabular*}
\end{table}

%% file: section/5_conclusion.tex
\section{Conclusion}
We presented MINT, which incorporates spatial transcriptomics supervision into pretrained pathology ViTs through a dedicated ST token and dual distillation. By decoupling molecular and morphological learning, MINT achieves the highest scores on both HEST-Bench and EVA in a backbone-agnostic manner, suggesting that cross-modal supervision offers a complementary axis beyond image-only self-supervised scaling.

 %% removed for anonymized MICCAI submission.
    
    % The following acknowledgement and disclaimer sections can be removed for the double-blind review process.  If and when your paper is accepted, reinsert the acknowledgement and the disclaimer clause in your final camera-ready version.
    % IF you opted to include the acknowledgement and disclaimer sections, they will count towards the 8-page limit.

% \begin{credits}
% \subsubsection{\ackname} A bold run-in heading in small font size at the end of the paper is
% used for general acknowledgments, for example: This study was funded

% \subsubsection{\discintname}
% It is now necessary to declare any competing interests or to specifically
% state that the authors have no competing interests. Please place the
% statement with a bold run-in heading in small font size beneath the
% (optional) acknowledgments\footnote{If EquinOCS, our proceedings submission
% system, is used, then the disclaimer can be provided directly in the system.},
% for example: The authors have no competing interests to declare that are
% relevant to the content of this article. Or: Author A has received research
% grants from Company W. Author B has received a speaker honorarium from
% Company X and owns stock in Company Y. Author C is a member of committee Z.
% \end{credits}